# An Evaluation of State-of-the-Art Large Language Models for Sarcasm Detection

**Juliann Zhou**
New York University, New York, NY 10002 USA

e-mail: kyz224@nyu.edu

**ABSTRACT** Sarcasm, as defined by Merriam-Webster, is the use of words by someone who means the opposite of what he is trying to say. In the field of sentimental analysis of Natural Language Processing, the ability to correctly identify sarcasm is necessary for understanding people's true opinions. Because the use of sarcasm is often context-based, previous research has used language representation models, such as Support Vector Machine (SVM) and Long Short-Term Memory (LSTM), to identify sarcasm with contextual-based information. Recent innovations in NLP have provided more possibilities for detecting sarcasm. In *BERT: Pre-training of Deep Bidirectional Transformers for Language Understanding*, Jacob Devlin et al. (2018) introduced a new language representation model and demonstrated higher precision in interpreting contextualized language. As proposed by Hazarika et al. (2018), CASCADE is a context-driven model that produces good results for detecting sarcasm. This study analyzes a Reddit corpus using these two state-of-the-art models and evaluates their performance against baseline models to find the ideal approach to sarcasm detection.

## 1. INTRODUCTION

Sarcasm refers to words that are opposite to their surface meanings. Commonly used in delivering sneering and mocking remarks, the word sarcasm comes from the ancient Greek word sarkázein, meaning "to tear flesh". It is sometimes used in a humorous or ambivalent manner and is largely context dependent. Identifying sarcasm is a challenging task in NLP, because the meaning of the words spoken opposes the speaker's true subjective opinion. The contextual nature of sarcasm makes detection even more difficult because it requires an analysis of the speaker's prior knowledge, tone, and intention.

Most computational models for detecting sarcasm rely on the content of the utterances in isolation. However, the speaker's intention may not be obvious without additional context. For example, the sentence 'I love solving math problems all weekend' may be sarcastic to many students but may not be so for a student who loves math. Hence, the context of a text is crucial for identifying sarcasms. This study investigated the use of contextual analysis using ContextuAl SarCasm Detector (CASCADE) and Bidirectional Encoder Representations from Transformers (BERT) for sarcasm detection. We conducted a qualitative analysis of the CASCADE model (Hazarika et al., 2018) and a BERT-based classifier model (Potamias et al., 2020) for sarcasm detection using a Reddit Corpus. Additionally, we will use BERT in conjunction with CASCADE to perform bidirectional training in our model with additional contextual information supplied by CASCADE and determine whether there is a significant performance improvement.

The language representation model BERT achieves state-of-the-art results for various NLP tasks (Devlin et al. 2018). BERT uses an innovative technique of applying bidirectional training to its transformer architecture in language modeling. The bidirectionally trained BERT model can grasp a deeper sense of language context and flow than single-direction language models, producing better results in NLP tasks, such as Question Answering, Natural Language Inference, and many others.

Recently, Hazarika et al. (2018) proposed a ContextuAl SarCasm Detector (CASCADE) model specifically designed for sarcasm detection. It utilizes a hybrid of content- and context-driven approaches. It processes contextual information by first profiling every author and every online community in their dataset before training and creates user embeddings of behavioral traits for sarcasm. It then performs concurrent training of the text and its surrounding discussions on the forum. Finally, CASCADE performs content modeling using a Convolutional Neural Network (CNN) to extract syntactic features of the text. The concatenated CNN representation with user embedding and discussion features is then used for classification.

Because sarcasm is most often dependent on its spoken context and BERT has achieved accurate results in other contextual analysis tasks, we want to apply the BERT model to sarcasm detection using a Reddit corpus. We also performed the same task using CASCADE, which is the current state-of-the-art model designed explicitly for contextual analysis in sarcasm detection. Finally, we conduct quantitative and qualitative analyses to compare the results of the two models. Additionally, we will investigate the possibility of using the two models in conjunction, augmenting BERT with additional contextual analyzing features supplied by CASCADE and determine if there is an improvement in performance.

## 2. RELATED WORK

Because of the prevalence of sarcasm in sentiment-bearing texts and the challenges in predicting sarcastic language with an automatic process, sarcasm detection has been a topic of interest in natural language processing research. In general, three types of approaches are used in sarcasm detection: rule-based approaches, content and contextual-based approaches, and deep learning- and transformer-based approaches.

Rule-based approaches identify sarcasm using rules that rely on the indicators of sarcasm. Reyes et al. (2012) first attempted to automate the identification of irony and sarcasm in social media. They used a rule-based approach that measured the unexpectedness factor in analyzing figurative language, including sarcasm. Barbieri et al. measured unexpectedness using the American National Corpus Frequency Data and the morphology of tweets with random forest (RF) and decision tree (DT) classifiers. Buschmeier et al. (2014) defined unexpectedness as an imbalance of emotions between words in a text.

Other rule-based approaches include the use of hashtag sentiment as a key identifier of sarcasm, which is frequently used in sarcastic tweets. In the study by Maynard and Greenwood (2014), sarcasm was detected by locating hashtags with sentiments that did not agree with the rest of the text. Reyes et al. (2012) used the likelihood of a simile being sarcastic as a rule and presented a 9-step approach where a simile is determined as non-sarcastic using the number of Google search results. Ghosh et al. (2015) identified sarcasm with support vector machines (SVM) to identify contradictions within a tweet as a measure of unexpectedness and achieved state-of-the-art results.

Content and contextual-based approaches rely on the dependency of sarcasm on context. It utilizes features that reveal information about the content, including acronyms and adverbs, N-gram patterns, statistical and semantic features, and semi-supervised attributes such as word frequencies. Irazú et al. (2018) used the affective and structural features of text with a conventional learning classifier to analyze sarcastic language contexts. They then conducted a follow-up study with a knowledge-based k-NN classifier and a feature set that captured both the structural and emotional linguistic phenomena in the context of sarcasm. Van Hee et al. (2015) achieved a significant improvement in LSTM deep neural network baseline results using a combination of lexical, semantic, and syntactic features with an SVM classifier. Through their study, Wallace et al. (2015) found that capturing contextual information by including previous and following comments on a Reddit corpus increased the accuracy of sarcasm detection. In a study by Rajadesingan et al. (2015), users' behavioral information was used as contextual information in sentiment analysis of tweets. Deep-learning approaches, such as recurrent neural networks and other language transformers, use word embeddings (the mapping of words to real-valued vectors) as the key factor in sarcasm detection. Ghosh et al. (2015) implemented a combination of word embeddings and convolutional neural networks (CNN). Kumar and Garg engineered a bidirectional LSTM model using an ensemble of shallow classifiers with lexical, pragmatic, and semantic features, and coupled the model with a CNN in a subsequent study by Kumar and Garg (2019). Recently, transformers, a specific class of deep learning models, have become prominent in sarcasm detection and natural language processing. The bidirectional encoder representation transformer (BERT) model has achieved state-of-the-art results in many NLP tasks. Khatri et al. (2020) used the BERT model in sarcasm detection by fine tuning an instance of BertForSequenceClassification and has shown significant improvement in performance compared to bidirectional LSTM and SVM sarcasm detection.

## 3. EVALUATION

### A. Objective

The objective of this study is to utilize both contextual- and deep-learning-based approaches in sarcasm detection, given their successes in previous studies, and to compare their performances. The first approach (Hazarika et al., 2018) utilized a contextual-based approach to analyze sarcastic online comments with users' personality, stylometrics, and discourse features. The second was a deep learning approach using the RoBERTa model (Potamias et al., 2020).

We implemented these two models, as they have previously achieved state-of-the-art results and are representative of contextual-based and deep learning-based approaches in the area of sarcasm detection. These two models will be used in the Reddit SARC corpus (Khodak et al., 2017) for sarcasm detection. Finally, we analyzed their performance and compared it with those of other baseline systems.

### B. Methodology

We applied two models, BERT and CASCADE, to test the performance of the transformer-based and contextual-based methods on the same Reddit corpus SARC. Both methods have shown success and are currently state-of-the-art in their respective approaches for sarcasm detection.

### 1) CASCADE

The CASCADE model (Hazarika et al., 2018) leverages content- and context-based information from a Reddit comment for classification. For content modeling, the comment was converted into a vector representation using a Convolutional Neural Network (CNN). Using a CNN, the model can extract location-invariant local patterns and generate abstract representations of the text, which encapsulates useful syntactic and semantic information for sarcasm detection.

The contextual modeling of CASCADE comprises the user embeddings and discourse features of all users and discussion forums. The contextual model generates user embeddings by modeling the user's stylometric and personality features and combining them with CCA for a single representation.

For the personality features, the user embeddings capture users' traits to determine their sarcasm tendencies by analyzing each user's accumulated historical posts. Hazarika et al. (2018) used a CNN pretrained on a benchmark corpus of essays labeled with Big Five personality traits. By iterating the overall comments, it was possible to identify the personality traits inferred from each comment. Finally, it uses the expectations of all comments made by the user as the user's overall personality feature vector.

For the stylometric feature, the user embeddings evaluate the writing style of the user. Hazarika et al. (2018) applied an unsupervised representation learning method, ParagraphVector, developed by Le and Markolov, which encodes a user's writing style and has been well-tested for sentiment classification tasks. Finally, the model combines both stylometric and personality features in the user embeddings with CCA (Canonical Correlation Analysis).

In addition, CASCADE incorporates a discourse feature that considers the contextual information provided by the surrounding discourse and the characteristics unique to the discussion forum section in which the comment is made. Hazarika et al. (2018) used a measure similar to ParagraphVector to generate feature vectors.

Finally, the user embeddings and discourse feature vectors are concatenated into a unified representation, which is projected onto the output layer of the CNN to obtain a softmax probability that classifies the comment as either sarcastic or nonsarcastic. Finally, Hazarika et al. (2018) calculated the categorical cross-entropy using probability estimates.

### 2) BERT

For the BERT model, we implemented RCNN-RoBERTa (Recurrent CNN Robustly Optimized BERT Approach) model (Potamias et al., 2020). The RoBERTa model optimized BERT with training from ten times more data and far more epochs. It also uses 8-times larger batch sizes and a byte-level BPE vocabulary instead of the character-level vocabulary used in BERT. RoBERTa also modified the single static mask used in BERT with dynamic masking techniques.

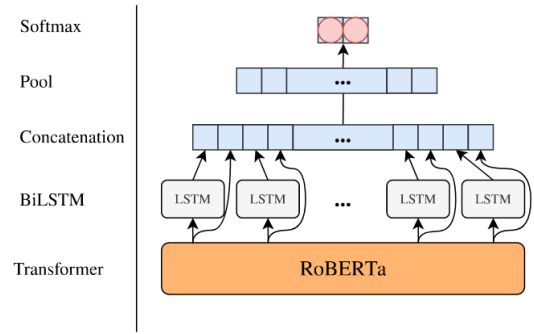

**FIGURE 1.** RCNN-RoBERTa transformer-based architecture proposed by Potamias et al. (2020)

The end-to-end model devised combines pretrained RoBERTa weights with an RCNN to capture contextual information. Identifying sarcasm within a sentence requires finding dependencies within RoBERT as pre-trained word embeddings. Potamias et al. (2020) applied an RNN layer that captures reliable temporal information instead of 1D convolution layers in vanilla RoBERTa, which cannot detect dependencies among word embeddings. The contradictions and long-time dependencies within a sentence are used as strong identifiers of sarcastic language. Because RNN's capturing of temporal relationships between words is strongly biased toward later words than previous ones, but unbiased CNNs are dependent on kernel sizes, Potamias et al. (2020) utilized an RCNN model to capture unbiased recurrent informative relationships within text.

A bidirectional LSTM layer was then applied and fed into RoBERT as the final hidden layer weight. Potamias et al. (2020) concatenated the output of LSTM with embedded weights and passed the result through a feedforward network, which acted as a 1D convolution layer with a large kernel and max-pooling layer. Finally, the output layer was processed using a softmax function.

*C. Dataset*

The SARC corpus (Khodak et al., 2017) includes 1.3 million remarks of self-annotated comments from the online forum Reddit. Each entry in the dataset contains the original comment sentence, response sentence, and annotation by the author of whether the comment is sarcastic. Reddit is organized into topic-specific discussion forums known as subreddits. In each subreddit, users comment on each other either on a titled post or on each other's comments under the titled post. This results in a tree-like comment structure, where the roots are topic-specific subreddits, followed by posts within a subreddit, comments made in posts on the subreddit, and comments on a post's other comments. The structure was then unraveled in a linear format, incorporating the author's details for each comment. The following is an example of a data point:

**Ancestor**: What will we call Bill Clinton if Hillary is elected president?

**Response**: I can think of a few names (**Label: Sarcastic**)

In our evaluation, we considered one variant of the SARC dataset, r/politics, a subreddit within the main Reddit forum that centers around politics. There were several reasons for this selection. The *r/politics* subreddit is a benchmark dataset (Khodak et al., 2017). It is also the dataset used for the evaluation of most baseline systems, with which we compare our results. Another reason is that sarcasm detection requires understanding of the background information being discussed. Even humans have trouble detecting sarcasm on unfamiliar topics, such as topics on obscure art and hobbies forums. Hence, this is the only subsample of the SSARC dataset evaluated by humans in the original tests conducted by Khodak et al. (2017) because most evaluators had sufficient background information. The SARC r/politics subsample consists of 17 thousand sequences, with an average proportion of sarcastic comments of 23.2%.

*D. Training Details*

For the CASCADE model, 20% of the training data were reserved for validation. Table 1 shows the distribution of the SARC Politics dataset for training and testing.

TABLE I
DISTRIBUTION OF SARC POLITICS FOR TRAINING AND TESTING

| Training set | | | | Testing set | | | |
|---|---|---|---|---|---|---|---|
| no. of comments | | avg. no. of words per comments | | no. of comments | | avg. no. of words per comments | |
| non-sarc | sarc | non-sarc | sarc | non-sarc | sarc | non-sarc | sarc |
| 6834 | 6834 | 64.74 | 62.36 | 1703 | 1703 | 62.99 | 62.14 |

Non-sarc: non-sarcastic, sarc: sarcastic

With the validation set, hyperparameter tuning was performed through RandomSearch and optimized using the Adam optimizer, as proposed by Kingma and Ba. Each comment was padded to a uniform length of 100 words for batch modeling in the CNN. The optimal hyperparameters were $\{ds, dp, dt, K\} = 100, dem = 300, ks = 2, M = 128$, and $\alpha = ReLU$.

For training with the RCNN-RoBERTa (Potamias et al., 2020), we implemented the same RandomSearch of the training data to find the following combinations of hyperparameters that produced the best results. The model was trained for five epochs and took approximately 3h to train.

*E. Evaluation Metrics*
The evaluation metric we will use for this paper is accuracy and F1 score.

$$\text{Accuracy} = \frac{\text{truePositive} + \text{trueNegative}}{\text{all}}$$

$$F1 = 2 \times \frac{\text{precision} \times \text{recall}}{\text{precision} + \text{recall}}$$

We chose accuracy as the primary evaluation metric because it is the evaluation metric used for most baseline models, which allows for easy comparisons. Because the dataset had a balanced number of data points in each class, class imbalance, such as a large number of actual negatives, will not be a problem. In this case, accuracy would be a good metric for a dataset with an even number of data points. However, we also include the F1 score as our secondary measure, which will guard against an uneven class distribution.

TABLE 2
RCNN-RoBERTa TRAINING HYPERPARAMETERS

| Hyperparameter | Value |
|---|---|
| RoBERTa layers | 12 |
| RoBERTa attention heads | 12 |
| LSTM units | 64 |
| LSTM dropout | 0.1 |
| Batch size | 10 |
| Adam Epsilon | 1e-6 |
| Epochs | 5 |
| Learning rate | 2e-5 |
| Weight decay | 1e-5 |
| RoBERTa layers | 12 |
| RoBERTa attention heads | 12 |
| LSTM units | 64 |
| LSTM dropout | 0.1 |
| Batch size | 10 |
| Adam Epsilon | 1e-6 |

*F. Baseline Models*
- **Average Human Performance:** This baseline of average human performance (Khodak et al., 2017) proposed the SARC dataset and asked human evaluators to classify sarcastic comments in the SARC politics dataset.
- **Bag of Words Baseline:** The model uses an SVM classifier with the input features of a comment's word count.
- **CNN-SVM:** This model (Poria et al, 2016) uses a CNN for content modeling and
- SVM for classification.
- **CUE-SVM:** This approach (Amir et al., 2016) utilizes a similar user embedding with *ParagraphVector* as that implemented in CASCADE, which is then fed into the CNN.

*G. Results*
Table 2 presents the performance results for the SARC politics dataset for the two models implemented in our study and the baseline models. The baseline models are presented in the previous section.

TABLE 3
EVALUATION RESULTS ON SARC POL

| Model | Accuracy | F1 |
|---|---|---|
| Average Human Performance | 0.82 | - |
| Bag of Words Baseline | 0.59 | 0.60 |
| CNN-SVM | 0.65 | 0.67 |
| CUE-SVM (Amir et al., 2016) | 0.69 | 0.70 |
| CASCADE | 0.74 | 0.75 |
| RCNN-RoBERTa | 0.79 | 0.78 |

The bag-of-words approach achieved the lowest performance, and most neural network approaches achieved higher performance, indicating that more sophisticated deep learning methods are better able to classify sarcasm than the word-targeting BoW model. As the SARC politics dataset is balanced, there is little significant difference between the accuracy and F1. Because CUE-SVM implemented similar user embeddings as the stylometric embeddings in CASCADE, we observed improved performance compared with a similar model without user embeddings, suggesting that contextual information on users improves sarcasm detection.

Both CASCADE and RCNN-RoBERTa outperformed the baseline models with statistical significance. Because the CASCADE model outperforms CUE-SVM, the personality and discourse features in the CASCADE model may account for this difference. Hence, contextual information on authors' personalities and surrounding discussions may contribute to sarcasm detection. RCNN-RoBERTas achieved better performance than all other methods, which indicates that implementing transformer architectures in deep learning methods of sarcasm detection could improve the performance of traditional deep learning approaches.

## 4. CONCLUSION

In this study, we implemented two different state-of-the-art models, CASCADE (Hazarika et al., 2018) and RCNN-RoBERTa (Potamias et al., 2020) to classify sarcastic comments in a Reddit corpus. CASCADE and RCNN-RoBERTa are representative contextual-based and deep-learning-based approaches used for sarcasm detection. We found that contextual information, such as user personality embeddings, could significantly improve performance, as well as the incorporation of a transformer RoBERTa, compared with a more traditional CNN approach. Given the success of both contextual- and transformer-based approaches, as shown in our results, augmenting a transformer with additional contextual information features may be an avenue for future experiments.